%% file: paper.tex
\documentclass[]{fairmeta}
\newcolumntype{C}{>{\centering\arraybackslash}X}

\usepackage{times}  % DO NOT CHANGE THIS
\usepackage{helvet}  % DO NOT CHANGE THIS
\usepackage{courier}  % DO NOT CHANGE THIS
\usepackage{algorithmic}
\usepackage[table]{xcolor} 
\usepackage{newfloat}
\usepackage{wrapfig}
\usepackage{tabularx}
\usepackage{textcomp}
\usepackage{stfloats}
\usepackage{url}
\usepackage{verbatim}
\usepackage{titlesec}
\usepackage{tocloft}
\usepackage{adjustbox}
\usepackage{pifont}
\usepackage{tikz}
\usepackage{comment}
\usepackage{amsmath,amssymb} % define this before the line numbering.
\usepackage{color}
\usepackage{natbib} 
\usepackage{hyperref}
\usepackage{graphicx}    % 用于插入图片
\usepackage{multirow} % 表格多行合并
\usepackage{booktabs} % 表格更好的线条
\usepackage{subcaption} % 支持 subtable 和 subfigure
\RequirePackage{xspace}
\makeatletter
\DeclareRobustCommand\onedot{\futurelet\@let@token\@onedot}
\def\@onedot{\ifx\@let@token.\else.\null\fi\xspace}
\usepackage[most]{tcolorbox}
\usepackage{xcolor}

\newcommand{\eg}{\textit{e.g.}\xspace} % 定义 \eg
\definecolor{Real}{HTML}{0072B2}
\definecolor{RoboSimGS}{HTML}{D55E00}

\definecolor{headerpurple}{HTML}{d8d2fc}

\newcommand{\our}{{RoboSimGS}\xspace}

\def\eg{\emph{e.g}\onedot}

\makeatother

\definecolor{adptorange}{RGB}{248, 205, 172}
\definecolor{cmpblue}{RGB}{189, 215, 238}
\definecolor{cmpblue}{RGB}{189, 215, 238}

\definecolor{our_red}{RGB}{232,157,160}
\definecolor{our_blue}{RGB}{136,206,230}
\definecolor{our_orange}{RGB}{246,200,168}
\definecolor{our_green}{RGB}{178,211,164}

\definecolor{attn_code0}{RGB}{247,215,200}
\definecolor{attn_code1}{RGB}{238,169,139}
\definecolor{mlp_code0}{RGB}{204,201,221}
\definecolor{mlp_code1}{RGB}{102,95,153}

\definecolor{token_blue}{RGB}{84, 120, 140}
\usepackage{pifont}       % \ding{xx}
\usepackage{bbding}       % \Checkmark,\XSolid,... (需要和pifont宏包共同使用)
\usepackage{fontawesome}
\usepackage{xspace}

\usepackage{float}

\newlength\savewidth

\newcolumntype{x}[1]{>{\centering\arraybackslash}p{#1pt}}
\newcolumntype{y}[1]{>{\raggedright\arraybackslash}p{#1pt}}
\newcolumntype{z}[1]{>{\raggedleft\arraybackslash}p{#1pt}}

\renewcommand{\paragraph}[1]{\vspace{1mm}\noindent\textbf{#1}}
\usepackage{colortbl}
\usepackage{xcolor}

\renewcommand{\paragraph}[1]{\vspace{1.25mm}\noindent\textbf{#1}}

\usepackage{algorithm}
\usepackage{listings}

\definecolor{codeblue}{rgb}{0.25, 0.5, 0.5}
\definecolor{codekw}{rgb}{0.35, 0.35, 0.75}
\lstdefinestyle{Pytorch}{
    language = Python,
    backgroundcolor = \color{white},
    basicstyle = \fontsize{9pt}{8pt}\selectfont\ttfamily\bfseries,
    columns = fullflexible,
    aboveskip=1pt,
    belowskip=1pt,
    breaklines = true,
    captionpos = b,
    commentstyle = \color{codeblue},
    keywordstyle = \color{codekw},
}

\definecolor{green}{HTML}{009000}
\definecolor{red}{HTML}{ea4335}

\title{High-Fidelity Simulated Data Generation for Real-World Zero-Shot Robotic Manipulation Learning with Gaussian Splatting}

\author[* 1, 2]{Haoyu Zhao}
\author[* 5]{Cheng Zeng}
\author[* 1]{Linghao Zhuang}
\author[2, 3]{Yaxi Zhao}
\author[2, 3]{Shengke Xue}
\author[6]{Hao Wang}
\author[2]{Xingyue Zhao}
\author[4]{Zhongyu Li}
\author[2, 3]{Kehan Li}
\author[\dagger 2, 3, 7]{Siteng Huang}
\author[2, 3]{Mingxiu Chen}
\author[2, 3]{Xin Li}
\author[2, 3]{Deli Zhao}
\author[\dagger 1]{Hua Zou}

\affiliation[1]{Wuhan University}
\affiliation[2]{DAMO Academy, Alibaba Group}
\affiliation[3]{Hupan Lab}
\affiliation[4]{The Chinese University of Hong Kong}
\affiliation[5]{Tsinghua University}
\affiliation[6]{Huazhong University of Science and Technology}
\affiliation[7]{Zhejiang University}

\contribution[*]{Equal contribution}
\contribution[\dagger]{Corresponding author}

\input{Section/0_abstract}
\homepage{\url{https://robosimgs.github.io/}\par\vspace{0.8\baselineskip}}
\date{\today}

\begin{document}
\thispagestyle{firstheader}
\maketitle
\pagestyle{empty}

\input{Section/1_introduction}

\input{Section/2_related_work}

\input{Section/3_method}
\input{Section/4_experiments}
\input{Section/5_conclusion}

\bibliographystyle{assets/plainnat}
\bibliography{paper}

\end{document}

%% file: Section/0_abstract.tex
\abstract{
The scalability of robotic learning is fundamentally bottlenecked by the significant cost and labor of real-world data collection. While simulated data offers a scalable alternative, it often fails to generalize to the real world due to significant gaps in visual appearance, physical properties, and object interactions. 
To address this, we propose \our, a novel Real2Sim2Real framework that converts multi-view real-world images into scalable, high-fidelity, and physically interactive simulation environments for robotic manipulation. Our approach reconstructs scenes using a hybrid representation: 3D Gaussian Splatting (3DGS) captures the photorealistic appearance of the environment, while mesh primitives for interactive objects ensure accurate physics simulation. Crucially, we pioneer the use of a Multi-modal Large Language Model (MLLM) to automate the creation of physically plausible, articulated assets. The MLLM analyzes visual data to infer not only physical properties (\eg, density, stiffness) but also complex kinematic structures (\eg, hinges, sliding rails) of objects.
We demonstrate that policies trained entirely on data generated by \our achieve successful zero-shot Sim2Real transfer across a diverse set of real-world manipulation tasks. Furthermore, data from \our significantly enhances the performance and generalization capabilities of SOTA methods. Our results validate \our as a powerful and scalable solution for bridging the Sim2Real gap.
}

%% file: Section/1_introduction.tex
\section{INTRODUCTION}
The pursuit of generalist robot policies capable of open-world manipulation~\citep{black2024pi0visionlanguageactionflowmodel,bjorck2025gr00t,team2025gemini,kim2024openvla} represents a grand ambition in Embodied AI and robotics. These end-to-end models learn directly from raw sensory input and promise capabilities like language instruction following, task transfer, and in-context learning. However, this ambition is fundamentally constrained by the data acquisition bottleneck. While teleoperation systems~\citep{li2025teleopbench,zhao2025smap,zhao2023learning,whitney2018ros} offer a partial solution for gathering expert demonstrations, they fails to overcome the fundamental bottleneck of high human effort~\citep{yu2025real2render2real}, thus limiting the scalability of models that promise language-guided control and generalization.

In contrast, generating the synthetic data under the simulated environment offers the advantage of exponential scalability with computational resources, making it an appealing and renewable alternative for training robotic policies. Those Sim2Real approaches suffer from the large domain gaps in geometric representation, visual appearance, and physical material behaviors between simulated and real-world environments, making transfer the learned policy into the real world remains challenging~\citep{huber2024domain,tobin2017domain}. Substantial efforts have been made to bridge this gap through techniques like domain randomization~\citep{sadeghi2016cad2rl,tobin2017domain} and system identification~\citep{chebotar2019closing,tan2018sim}. They enhance the agent’s robustness by simulating real-world noises and aligning the agent dynamic model with the real-world settings. While these methods have yielded certain advancements, their effectiveness is fundamentally limited by the simulator’s capabilities. Traditional simulators often fail to provide visually realistic observations, dynamic interactions, and diverse environmental variations, limiting the agent’s capacity to generalize effectively beyond the simulation environments. 

To directly tackle the source of the reality gap, the Real2Sim2Real (R2S2R) paradigm has recently emerged as a transformative approach~\citep{li2024robogsim,yang2025novel,han2025re,wu2024rl,yu2025real2render2real}. The core insight is to perform Real2Sim reconstruction via radiance field methods, such as NeRF~\citep{mildenhall2021nerf} and 3D Gaussian Splatting (3DGS)~\citep{kerbl20233d}, and insert learned photorealistic representations into the simulator to drastically reduce the domain gap. Pioneering works like Robo-GS~\citep{lou2024robo} have demonstrated the potential of this R2S2R pipeline, introducing a hybrid representation to generate digital assets enabling high-fidelity simulation. However, despite achieving promising visual fidelity, these frameworks fall short in a critical dimension: physical interaction. Their focus on photorealism results in a world that is largely static and non-interactive, limiting their application to observational tasks and preventing the training of policies for complex, contact-rich manipulation.

To tackle these issues, we propose \textbf{\our}, a novel R2S2R framework designed to bridge this crucial gap between photorealism and physical interactivity. At its core, \our constructs a hybrid scene representation, synergizing the photorealism of 3DGS for static backgrounds with the explicit geometry of mesh primitives for interactive objects, which allows for both high-fidelity visuals and corresponding physical collision. Crucially, to bring these objects to life, we pioneer the use of a Multi-modal Large Language Model (MLLM) to directly infer an object's physical properties (\eg, density and stiffness) and kinematic structure (\eg, hinges, drawers) from multi-view images, transforming a static scene into a dynamic, interactive ``sandbox''. We further enhance policy robustness through holistic scene augmentation that randomizes object properties, camera views, lighting, and trajectories. Extensive experiments demonstrate that policies trained exclusively on our simulated data achieve remarkable zero-shot transfer to the real world across a variety of manipulation tasks. Furthermore, \textit{augmenting limited real-world data with our generated data significantly boosts the performance and generalization capabilities of state-of-the-art visuomotor policies}, highlighting \our's value as a scalable and effective data generation solution. We summarize our contributions as follows:

\begin{itemize}
    \item We introduce a novel R2S2R framework that generates photorealistic and physically interactive simulation environments from real-world scenes, built upon a hybrid (3DGS + mesh) representation.
    \item We pioneer the use of MLLM to automatically create physically-plausible and articulated assets.
    \item Extensive experiments demonstrate that our simulated data not only enables remarkable zero-shot Sim2Real transfer, but also significantly boosts the performance and generalization of existing SOTA models.
\end{itemize}

%% file: Section/2_related_work.tex
\section{RELATED WORK}
\begin{figure*}[t]
  \centering
  \includegraphics[width=\linewidth]{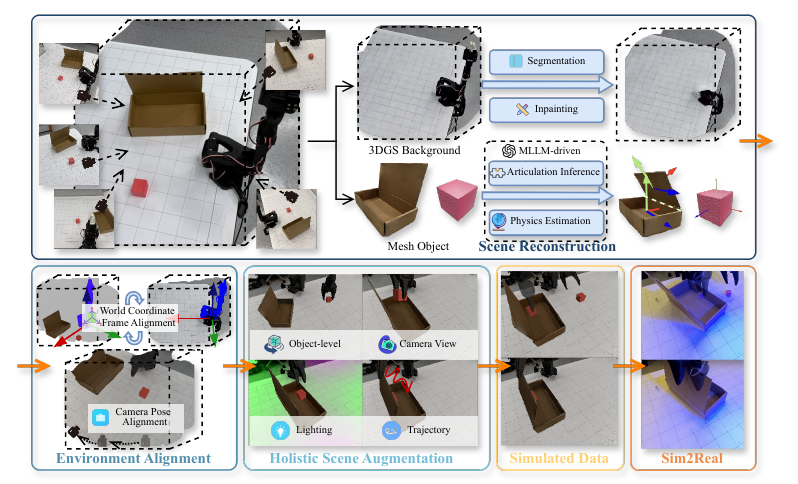}
  \caption{\textbf{Pipeline of \our}. Starting from multi-view images, we first perform Scene Reconstruction to create a hybrid representation with a photorealistic 3DGS background and interactive mesh objects. A key step involves using a Multi-modal Large Language Model (MLLM) for automatic Physics Estimation and Articulation Inference. The scene is then aligned with the simulator with Sim2Real Environment Alignment. Finally, we apply Holistic Scene Augmentation to generate diverse simulated data. Policies trained on this data can be deployed directly to the real world.}
  \label{fig:pipeline}
\end{figure*}

\subsection{Data Generation for Robotics}
Addressing data scarcity is a critical challenge in robotics. A primary strategy is computational generation of new data from limited source demonstrations. These approaches can be broadly categorized into two main branches: trajectory-level generation and observation-level augmentation.

One prominent line of work focuses on generating new action trajectories~\citep{jiang2025dexmimicgen,xue2025demogen,mandlekar2023mimicgen}. These methods primarily operate on a principle of relative-pose transformation. At their core, they exploit SE(3)-equivariance to replay transformed versions of a source trajectory under new object poses. While effective for pose variation, their reliance on a rigid, object-level formulation means they fundamentally struggle to generalize to novel object geometries, a failure point for tasks requiring precise manipulation.

Another line of work focuses on augmenting the visual appearance of observations to improve the robustness of visuomotor policies~\citep{hansen2021generalization,laskin2020reinforcement,yu2023scaling}. These methods are achieved by employing generative models, such as for inpainting robot embodiments or other scene elements into 2D images. While 2D image-space augmentations are computationally efficient and easy to implement, their inherent lack of 3D spatial awareness is a critical drawback. Operating directly on pixels without geometric context, these methods can generate unrealistic augmentations—such as objects pasted at physically impossible locations or textures that ignore object boundaries—which limits their effectiveness for tasks grounded in 3D interaction. We propose \our, a Real2Sim2Real framework that addresses data scarcity by creating a photorealistic and physically interactive simulator. Using a hybrid 3D representation (3DGS + mesh), \our generates rich, physically-plausible synthetic data at scale. This approach overcomes the core limitations of augmenting 2D observations or trajectories, providing inherent 3D spatial awareness and robust generalization to novel object geometries and interactions.

\subsection{Sim2Real Transfer in Robotics}
To improve the feasibility of deploying models in the real world, many researchers strive to bridge the gap between simulation and reality with Sim2Real methods~\citep{mouret201720}. Previous Sim2Real methods can be broadly classified
into two categories: domain randomization~\citep{huber2024domain}, domain adaptation~\citep{bousmalis2018using}. Domain randomization methods are designed to expand the operational envelope of a robot in a simulator by introducing randomness, which involves varying task-related parameters in the simulation to cover a broad range of real-world conditions~\citep{exarchos2021policy,huber2024domain,andrychowicz2020learning}. Domain adaptation approaches aim to extract a unified feature space of simulated and real environments, facilitating the training and migration within the unified feature space~\citep{bousmalis2017unsupervised,long2015learning}. 
These methods introduce the disturbances into the simulated environment, in which the policy of robots is learned. It develops the capacity to operate effectively in the real world with noise and unpredictability~\citep{wang2020reinforcement}. 

To tackle the reality gap at its source, the Real2Sim2Real (R2S2R) paradigm has emerged, aiming to automatically reconstruct high-fidelity digital twins from real-world scene to build simulated environments for policy learning~\citep{geng2025roboverse,jiang2025dexmimicgen}. However, these digital twins often fail to provide fully realistic observations, plausible dynamic interactions. For example, DexMimicGen~\citep{jiang2025dexmimicgen} uses fixed simulation assets. RoboVerse~\citep{geng2025roboverse} supports only rigid objects.

More recently, another line of work has adopted the R2S2R pipeline, utilizing 3D Gaussian Splatting (3DGS) to reconstruct photorealistic scenes~\citep{qureshi2024splatsim,li2024robogsim,yang2025novel,han2025re,lou2024robo}. 
For instance, SplatSim~\citep{qureshi2024splatsim} can create visually stunning, near-indistinguishable reconstructions of real scenes. 
However, despite their visual fidelity, these state-of-the-art R2S2R methods face a new, critical limitation: Interactions are typically limited to pre-defined, rigid assets, precluding the simulation of complex dynamics, articulations (like drawers or hinges), or non-rigid objects. In essence, they produce photorealistic but static ``digital snapshots'', not interactive robotic playgrounds. To bridge the physical interactivity gap, our method, RoboSimGS, creates interactive simulators through two key innovations. First, it uses a hybrid representation (3DGS + mesh) to separate static backgrounds from interactable objects. Second, and more critically, it pioneers the use of a Multi-modal Large Language Model (MLLM) to automatically infer material properties and kinematic constraints (\eg, hinges, sliding rails) directly from visual data. This automated generation of articulated assets enables rich, dynamic data creation at scale, facilitating zero-shot sim-to-real transfer for complex manipulation tasks.

%% file: Section/3_method.tex
\section{Methodlogy}
To bridge the Sim2Real gap, we introduce \our, a method that generates realistic and physically interactive simulation environments from multi-view images. Our approach is a two-stage pipeline designed to create a high-fidelity digital twin of a real-world scene. The first stage, Scene Reconstruction in Section.~\ref{sec:scence_reconstrcution}, builds a hybrid representation with 3D Gaussian Splatting (3DGS)~\citep{kerbl20233d} capturing the photorealistic appearance of the static environment, while explicit mesh for interactive objects. In the second stage, Sim2Real Environment Alignment in Section.~\ref{sec:s2r_alignment}, these components are seamlessly imported into a physics simulator, where the 3DGS model serves as the visual background and the meshes become dynamic, physics-enabled assets. We further enrich this environment via Holistic Scene Augmentation in Section.~\ref{sec:augmentation} to randomize objects, cameras, lighting, and trajectory to create diverse simulated data, as shown in Fig.~\ref{fig:pipeline}. A policy trained exclusively on this augmented synthetic data can achieve remarkable zero-shot transfer to real-world scenarios.

\subsection{Scene Reconstruction}
\label{sec:scence_reconstrcution}
While physics-based simulators offer high-fidelity physics, they struggle to render scenes with high-fidelity visual appearance, leaving a large gap from real world. Recent advancements in 3D reconstruction~\citep{kerbl20233d} provide high-quality rendering while keeping real-time rendering, promising to bridge these gaps. However, 3DGS~\citep{kerbl20233d} are not directly compatible with physics engines, as robustly converting them into interactive mesh formats remains a major research challenge~\citep{guedon2024sugar}. Therefore, we adopt a decoupled reconstruction approach. We use 3DGS for the static background to maximize visual fidelity and explicit meshes for interactive objects to ensure physical plausibility. The robot itself is excluded from this process, as it is readily available in the standard robot description file (URDF). Our process for each task-specific scene requires approximately 10 minutes of manual scanning to capture the necessary multi-view data.

While physics-based simulators offer high-fidelity physics, they struggle with photorealistic rendering, leaving a large visual gap from the real world. Recent advancements in 3D reconstruction like 3DGS~\citep{kerbl20233d} promise to bridge this gap with high-quality, real-time rendering. However, 3DGS is not directly compatible with physics engines, as robustly converting it into interactive mesh formats remains a major research challenge~\citep{guedon2024sugar}. To address this, our process begins with a manual scan of each task-specific scene, requiring approximately 10 minutes to capture the necessary multi-view data. We then adopt a decoupled reconstruction approach, using 3DGS for the static background to maximize visual fidelity and explicit meshes for interactive objects to ensure physical plausibility. The robot itself is excluded from this process, as it is readily available in a standard robot description file (URDF).

\subsubsection{Background Reconstruction}
\label{sec:env}
Background reconstruction contains most parts of the scenes and applies more photorealistic rendering using 3D Gaussian Splatting (3DGS)~\citep{kerbl20233d}. The scene is modeled by a set of 3D Gaussians, each defined by a position, opacity, color (via Spherical Harmonics), and a covariance matrix $\Sigma$. To render a novel view, these Gaussians are sorted by depth and projected onto the image plane. The final color $\mathbf{C}$ for each pixel is then synthesized via alpha-blending according to the volume rendering formula. This approach provides high-quality, real-time rendering. To enable precise spatial alignment between our reconstructed scene and the scene in the simulator, we append a learnable semantic feature vector $\mathbf{f}_i \in \mathbb{R}^d$ to each Gaussian~\citep{qiu2024feature,zhao2024sg}. The training of these semantic features is guided by language-driven supervision~\citep{radford2021learning}. We first define a set of text prompts corresponding to all semantic classes of interest (\eg, "a robot arm", "a red block", "a banana"). These prompts are then encoded into target text embeddings using the frozen CLIP text encoder. The dimension $d$ of our features is set to match that of the CLIP embeddings. For a given camera view, we render a 2D feature map $\hat{\mathbf{F}}$ and supervise it using a multi-class 2D ground-truth semantic mask. Specifically, we minimize a contrastive loss between the rendered feature for each pixel and the corresponding target text embedding.

The entire learning process is underpinned by the differentiable nature of 3DGS. The rendering process begins by culling outside the camera frustum~\citep{kerbl20233d}. The remaining Gaussians are then projected onto the 2D image plane using the projection matrix \(\mathbf{W}\) of the camera. This projection induces the following transformation on the covariance matrix $\mathbf{\Sigma}$:%
\begin{equation}
    \mathbf{\Sigma}^{'} = \mathbf{J}\,\mathbf{W}\,\mathbf{\Sigma}\,\mathbf{W^\top J^\top}\,,
\end{equation}
where $\mathbf{J}$ is the Jacobian of the projection matrix \(\mathbf{W}\). We can then render both the color and the visual features with the splatting algorithm:
\begin{equation}
    \{\hat{\mathbf{F}}, \hat{\mathbf{C}\}} = \sum_{i \in N} \{\mathbf{f}_i, \mathbf{c}_i\}\cdot \alpha_i\, \prod_{j = 1}^{i - 1} (1 - \alpha_j)\,,
\end{equation}%
where $\alpha_i$ is the opacity of the Gaussian conditioned on $\mathbf{\Sigma}^{'}$ and the indices $i\in N$ are in the ascending order determined by their distance to the camera origin.

\subsubsection{Object Reconstruction}
\label{sec:object_reconstruction}
For foreground objects, we employ ARCode~\citep{ARCode2022}, which can automatically segment the object, balancing usability and quality. However, the resulting mesh is inherently static, limiting its use in physical simulations. To support more complex manipulation tasks, we introduce a pipeline to automatically endow these static meshes with physically-plausible, articulated structures.

% ########################################################################
% ########################################################################
\begin{figure}[t]
  \centering
  \includegraphics[width=\linewidth]{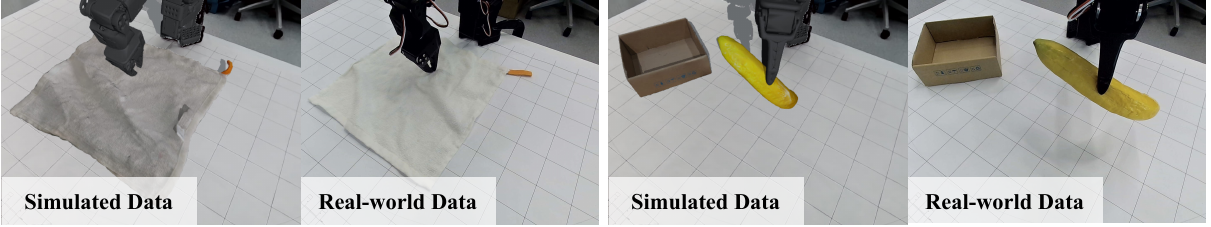} % <--- 修改这里
  \caption{\textbf{Qualitative comparison} between simulated data from \our (left) and real-world data (right).}
  \label{fig:sim_real}
\end{figure}
% ########################################################################
% ########################################################################

\medskip
\noindent
\textbf{MLLM-driven Articulation Inference.} 
Our process begins by employing a Multi-modal Large Language Model (MLLM), such as GPT-4o~\citep{achiam2023gpt}, to infer the object's kinematic structure. By providing the MLLM with multi-view renderings of the reconstructed mesh, it identifies the object's category and proposes its potential articulation, including the joint type (\eg, prismatic or revolute) and the semantic labels of the parts to be separated, such as \texttt{"drawer body"} and \texttt{"main cabinet"}. To partition the mesh according to the MLLM's proposal, we directly employ the open-vocabulary segmentation method from AffordDex~\citep{zhao2025towards}. This allows us to segment the object using the generated textual labels as direct prompts. Finally, with the mesh partitioned into a static base and a mobile part, the MLLM is again prompted to determine the precise joint parameters, such as its axis and motion limits. This information is used to automatically define a URDF-compatible joint, completing the creation of a fully interactive, articulated object for our simulation environment.

\medskip
\noindent
\textbf{MLLM-driven Physics Estimation.} 
Inferring an object's physical properties from its visual appearance is a long-standing challenge, as materials with similar aesthetics can exhibit vastly different physical behaviors.
While humans can perform this task through contextual reasoning, recent advances in Multi-modal Large Language Models (MLLMs) have demonstrated analogous capabilities in complex reasoning~\citep{driess2023palm,zhao2024automated}.
Inspired by this progress, we introduce a \textit{physics expert agent}, powered by GPT-4o~\citep{achiam2023gpt}, to automate the estimation of material properties.
Our agent processes a set of four orthographic views of a 3D asset to estimate its fundamental physical parameters: density ($\rho$ in kg/m\textsuperscript{3}), Young's modulus ($E$ in Pa), and Poisson's ratio ($\nu$, dimensionless).
These parameters are critical for high-fidelity physical simulation, as they govern the object's mass distribution, stiffness, and deformation response under external forces.

\begin{figure*}[t]
  \centering
  \includegraphics[width=\linewidth]{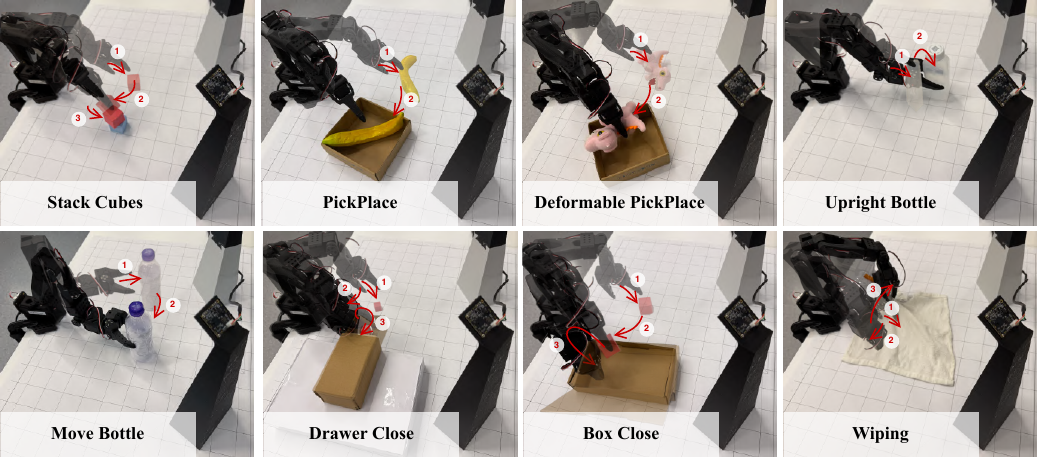}
  \caption{\textbf{Task illustration}. We design eight manipulation tasks for real-world evaluation: \textit{Stack Cubes}, \textit{PickPlace}, \textit{Deformable PickPlace}, \textit{Upright Bottle}, \textit{Move Bottle}, \textit{Drawer Close}, \textit{Box Close}, \textit{Wiping}, whose details are shown in Section.~\ref{sec:tasks}.}
  \label{fig:tasks_illustration}
\end{figure*}

\subsection{Sim2Real Environment Alignment}
\label{sec:s2r_alignment}
\subsubsection{World Coordinate Frame Alignment}
This step aligns the coordinate system of the 3DGS reconstruction described in Section.~\ref{sec:env} with that of the simulator, using the robot's geometry as a common anchor. For this process, the robots in both scenes are fixed at a default joint configuration. The alignment thus reduces to finding the rigid transformation between the robot's URDF base frame $\mathcal{R}_{urdf}$ and the 3DGS world frame $\mathcal{R}_{gs}$. To compute this transformation matrix, we first generate a point cloud of the robot's geometry from both the URDF model and the 3DGS reconstruction. We then apply the Iterative Closest Point (ICP)~\citep{besl1992method} algorithm, which finds the optimal rigid transformation $\mathcal{T}_{\text{scene}}$ by solving the following minimization problem:
\begin{equation}
    \mathcal{T}_{\text{scene}} = \underset{R \in SO(3), \mathbf{t} \in \mathbb{R}^3}{\arg\min} \sum_{i} \| (R \mathbf{p}_i^{\text{gs}} + \mathbf{t}) - \mathbf{q}_i^{\text{urdf}} \|^2,
\end{equation}
where $\{\mathbf{p}_i^{\text{gs}}\}$ and $\{\mathbf{q}_i^{\text{urdf}}\}$ are the point clouds from the 3DGS reconstruction and the URDF, respectively, and the transformation consists of a rotation $R \in SO(3)$ and a translation $\mathbf{t} \in \mathbb{R}^3$. By applying the resulting matrix $\mathcal{T}_{\text{scene}}$, we map all elements from the 3DGS scene into the simulator's coordinate system to get a aligned environment.

\subsubsection{Camera Pose Alignment}
\label{sec:cam_align}
To precisely align the simulated camera with its real-world counterpart, we formulate the problem as an optimization task that minimizes the photometric error between rendered and real images. Given a real-world reference image $I_{real}$, our objective is to solve for the optimal camera pose $\mathcal{T}_{cam}$, which comprises its position and rotation. We minimize the following loss function:
\begin{equation}
    \mathcal{L}_{cam} = \| \mathcal{R}(\mathcal{T}_{cam}) - I_{real} \|.
\end{equation}
where $\mathcal{R}(\cdot)$ is our differentiable rendering function that synthesizes an image from a given camera pose, and $ \| \cdot \|$ denote the L1 norm.

Starting from a rough initial pose $\mathcal{T}_{cam}^{init}$, we leverage the differentiability of the 3DGS renderer to compute the gradient of the loss with respect to the camera parameters. The camera pose is then iteratively refined using gradient descent until convergencee, resulting in a simulated camera pose that is precisely aligned with the real-world view.

With both the world coordinate frame and camera pose aligned, our hybrid representation is now fully integrated into the simulator, creating a high-fidelity digital twin ready for large-scale data generation, as shown in Fig.~\ref{fig:sim_real}.

\subsection{Holistic Scene Augmentation}
\label{sec:augmentation}
While the previous steps create a visually and physically faithful digital twin, a static environment is insufficient for training a robust policy that can handle real-world variability. To bridge this gap, we introduce Holistic Scene Augmentation, a process that systematically randomizes multiple facets of the simulated environment to create a diverse dataset.

\subsubsection{Object-level Augmentation}
To enhance the policy's robustness to variations in object placement and size, we perform object-level augmentation. We randomize the 6-DoF pose (position and orientation) and uniform scale of interactive objects, each sampled from a predefined range. This process ensures the model is exposed to a diverse set of object configurations during training, thereby improving its ability to generalize to unstructured real-world environments.

\subsubsection{Camera View Augmentation}
A key advantage of 3D Gaussian Splatting (3DGS) is its inherent capability for high-fidelity novel view synthesis. To generate diverse yet realistic camera perspectives, we anchor our augmentation to the deployment environment. We begin with the camera extrinsics that were optimized to match the real-world setup, as detailed in Section~\ref{sec:cam_align}. We then apply random perturbations to this base pose by introducing random translations and rotations. This creates a distribution of views centered on the deployment perspective, training the policy to be robust against minor shifts in camera placement.

\subsubsection{Lighting Condition Augmentation}
Discrepancies in lighting conditions between training and deployment environments pose a significant challenge to policy generalization. To address this, we directly augment the visual attributes of 3D Gaussians through a combination of random scaling, offset, and noise to simulate global changes in color contrast and overall brightness. In contrast, random Gaussian noise is sampled independently for each Gaussian to mimic the sensor noise characteristic of real-world cameras. This approach creates robust training data that enables the policy to operate effectively under a wide variety of previously unseen lighting conditions.

\subsubsection{Trajectory Augmentation}
To further diversify the demonstration data and prevent the policy from overfitting to a single kinematic path, we introduce trajectory-level augmentation. Instead of generating a motion to a target, the robot's end-effector is first commanded to a randomized intermediate via-point using an inverse kinematics (IK) solver. This via-point is generated by applying a small positional offset to the final goal. This two-stage motion strategy breaks trajectory determinism and exposes the policy to a richer distribution of paths, enhancing its robustness and ability to recover from minor execution errors.

\begin{table*}[t]
 \centering
 \scriptsize
 \caption{\textbf{Performance improvement from augmenting real data with simulated data}. Adding just 50 synthetic demonstrations from \our provides a substantial performance boost across all evaluated methods.}
 \label{tab:exp}
 \begin{tabularx}{\textwidth}{lcCCCC}
 \toprule
\rowcolor{headerpurple!60}
 \textbf{Method} & \textbf{Data Source} & \textbf{Stack Cubes} & \textbf{PickPlace} & \textbf{Deformable PickPlace} & \textbf{Upright Bottle} \\ 
 \midrule
 \rowcolor{gray!10} DP~\citep{chi2023diffusion} & 50 Real & 0.60 & 0.71 & 0.77 & 0.86 \\
 DP~\citep{chi2023diffusion} & 50 \our & 0.54 & 0.60 & 0.69 & 0.66 \\
 \rowcolor{gray!10}DP~\citep{chi2023diffusion} & 100 \our & 0.57 & \textbf{0.83} & \textbf{0.86} & 0.82 \\
 \rowcolor[HTML]{D7F6FF}DP~\citep{chi2023diffusion} & 50 Real + 50 \our & \textbf{0.69} & \textbf{0.83} & \textbf{0.86} & \textbf{0.91} \\
 \hline
 \rowcolor{gray!10}$\pi_0$~\citep{black2024pi0visionlanguageactionflowmodel} & 50 Real & 0.40 & \textbf{0.94} & \textbf{0.97} & 0.63 \\
 $\pi_0$~\citep{black2024pi0visionlanguageactionflowmodel} & 50 \our & 0.34 & 0.80 & 0.83 & 0.57 \\
 \rowcolor{gray!10}$\pi_0$~\citep{black2024pi0visionlanguageactionflowmodel} & 100 \our & \textbf{0.54} & 0.91 & 0.94 & 0.69 \\
 \rowcolor[HTML]{D7F6FF}$\pi_0$~\citep{black2024pi0visionlanguageactionflowmodel} & 50 Real + 50 \our & \textbf{0.54} & \textbf{0.94} & 0.94 & \textbf{0.74} \\
 \hline
 \rowcolor{gray!10}DP~\citep{chi2023diffusion} & 50 \our w/o Physics Estimation & 0.54 & 0.60 & 0.54 & 0.57 \\
 DP~\citep{chi2023diffusion} & 50 \our w/o Holistic Scene Augmentation & 0.37 & 0.46 & 0.51 & 0.43 \\
 
 \midrule[1.5pt]
 
 \textbf{Method} & \textbf{Data Source} & \textbf{Move Bottle} & \textbf{Drawer Close} & \textbf{Box Close} & \textbf{Wiping} \\
 \midrule
 \rowcolor{gray!10} DP~\citep{chi2023diffusion} & 50 Real & 0.89 & 0.51 & 0.57 & 0.91 \\
 DP~\citep{chi2023diffusion} & 50 \our & 0.80 & 0.43 & 0.54 & 0.85 \\
 \rowcolor{gray!10}DP~\citep{chi2023diffusion} & 100 \our & \textbf{0.91} & 0.49 & 0.63 & 0.91 \\
\rowcolor[HTML]{D7F6FF} DP~\citep{chi2023diffusion} & 50 Real + 50 \our & \textbf{0.91} & \textbf{0.60} & \textbf{0.66} & \textbf{0.94} \\
 \hline
 \rowcolor{gray!10}$\pi_0$~\citep{black2024pi0visionlanguageactionflowmodel} & 50 Real & 0.37 & 0.49 & 0.54 & 0.00 \\
 $\pi_0$~\citep{black2024pi0visionlanguageactionflowmodel} & 50 \our & 0.51 & 0.57 & 0.54 & 0.69 \\
 \rowcolor{gray!10}$\pi_0$~\citep{black2024pi0visionlanguageactionflowmodel} & 100 \our & 0.54 & \textbf{0.60} & 0.63 & 0.88 \\
 \rowcolor[HTML]{D7F6FF}$\pi_0$~\citep{black2024pi0visionlanguageactionflowmodel} & 50 Real + 50 \our & \textbf{0.57} & \textbf{0.60} & \textbf{0.69} & 0.86 \\
\hline
 \rowcolor{gray!10}DP~\citep{chi2023diffusion} & 50 \our w/o Physics Estimation & 0.77 & 0.46 & 0.49 & 0.51 \\
 DP~\citep{chi2023diffusion} & 50 \our w/o Holistic Scene Augmentation & 0.54 & 0.29 & 0.34 & 0.71 \\
 \bottomrule
 \end{tabularx}
\end{table*}

\begin{table*}[t]
 \centering
 \scriptsize
 \setlength{\tabcolsep}{4pt} % 适当增加列间距
 \caption{\textbf{Performance when facing changing settings}. We report the success rate of DP~\citep{chi2023diffusion} under various settings. The policies trained with demonstrations generated by \our significantly boosts the policy's generalization capabilities.}
 \begin{tabular}{c c *{4}{>{\centering\arraybackslash}m{2.5cm}}}
 \toprule
 \rowcolor{headerpurple!60}
  \textbf{Data Source}
 & \textbf{Lighting Condition}
 & \textbf{Object Size}
 & \textbf{Scene Clutter}
 & \textbf{Camera Pose}
 & \textbf{Desktop Appearance} \\
 \midrule
   \rowcolor{gray!10} 50 Real & 0.00 & 0.06 & 0.00 & 0.23 & 0.00 \\
    % 100 Real & 0.0 & 0.0 & 0.0 & 0.0 & 0.0 \\
   50 \our & 0.46 & 0.63 & 0.51 & 0.57 & 0.49 \\
   % 100 \our & & & & & \\
  \rowcolor[HTML]{D7F6FF} 50 Real + 50 \our & \textbf{0.54} & \textbf{0.71} & \textbf{0.60} & \textbf{0.71} & \textbf{0.66} \\
  % 50 Real + 100 \our & & & & & \\
 \bottomrule
 \end{tabular}
 \label{tab:exp1}
\end{table*}

\begin{figure*}[t]
  \centering
  \includegraphics[width=0.8\linewidth]{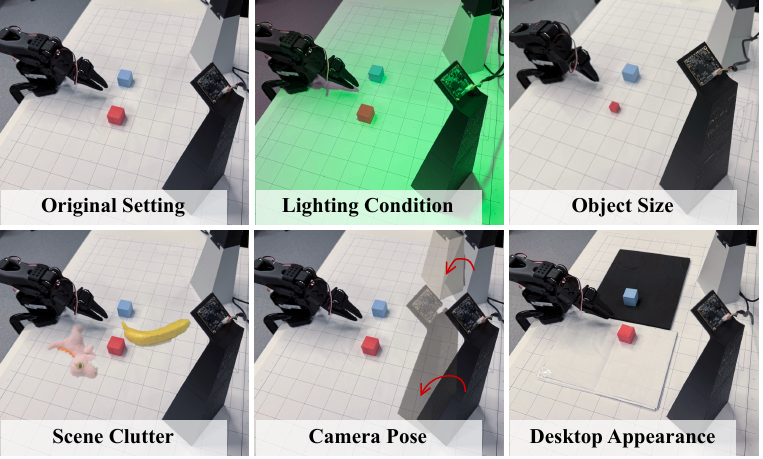}
  \caption{\textbf{Visualization} of policy performance under four challenging generalization settings designed to test robustness.}
  \label{fig:challenging_settings}
\end{figure*}

\begin{figure*}[t]
  \centering
  \includegraphics[width=\linewidth]{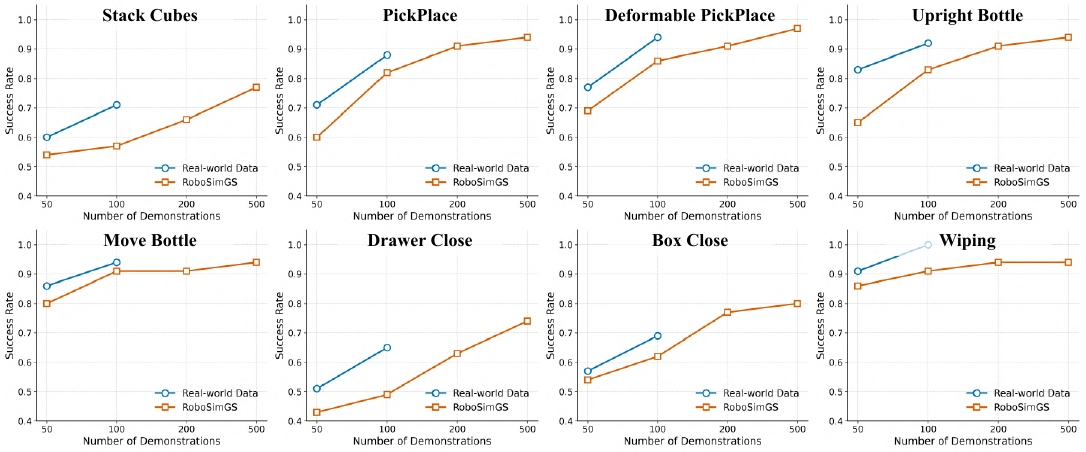}
  \caption{\textbf{Data scaling analysis} for Diffusion Policy~\citep{chi2023diffusion} on the \textit{Stack Cubes} task. The plot compares the success rate of policies trained on varying amounts of \textbf{\textcolor{Real}{real-world data}} versus purely \textbf{\textcolor{RoboSimGS}{simulated data generated by \our}}. Notably, the policy trained on 200 simulated demonstrations achieves a success rate comparable to one trained on 100 real-world demonstrations, highlighting the high quality and data efficiency of our generated data.}
  \label{fig:scaling}
\end{figure*}

%% file: Section/4_experiments.tex
\section{EXPERIMENTS}
\subsection{Experimental Setups}
\subsubsection{Hardware platform} 
We conduct our experiments across both real-world and simulated environments. For real-world data collection and policy evaluation, we utilize the LeRobot~\citep{cadene2024lerobot} framework with two cameras providing RGB observations which captures RGB observations at 640x480 resolution. Meanwhile, all simulation data is generated on a separate workstation configured with an Intel Core i5-14400F CPU and an NVIDIA RTX 5060 Ti GPU. 

\subsubsection{Policy and VLA model training details} 
To validate our simulated data fidelity, we evaluate several state-of-the-art policies: \textbf{Single-task methods}, such as Diffusion Policy~\citep{chi2023diffusion} across diverse tasks, and \textbf{Generalist VLA model}: $\pi_0$~\citep{black2024pi0visionlanguageactionflowmodel}. Policy inference is carried out on an NVIDIA H20 GPU.

\subsubsection{Metric}
The primary metric for evaluation is the Success Rate, defined as the percentage of successful task completions over 35 consecutive trials in the real world. A trial is deemed successful if the robot correctly finish the task in under 20 seconds.

\subsubsection{Robot tasks}
\label{sec:tasks}
To demonstrate the generalization ability of our pipeline across diverse manipulation challenges, we evaluate our method on eight distinct tasks, as shown in Fig.~\ref{fig:tasks_illustration}. These tasks cover a spectrum of skills, including stacking, articulated object interaction, deformable object handling, and tool use. The specific tasks are as follows: 
\begin{itemize}
    \item \textbf{\textit{Stack Cubes:}} Stacking a 3.5\,cm red cube onto a 3.5\,cm blue cube.
    \item \textbf{\textit{Pick \& Place:}} Picking a banana and placing it into a box.
    \item \textbf{\textit{Deformable Pick \& Place:}} A variation of the previous task, where the rigid banana is replaced with a soft toy to test deformable object manipulation.
    \item \textbf{\textit{Upright Bottle:}} Restoring an overturned bottle to its stable, upright orientation.
    \item \textbf{\textit{Move Bottle:}} Grasping and dragging a bottle across the workspace.
    \item \textbf{\textit{Drawer Close:}} A long-horizon task requiring the robot to place a 2\,cm cube into a drawer and then close it.
    \item \textbf{\textit{Box Close:}} A similar long-horizon task involving placing a 3.5\,cm red cube into a box and closing its lid.
    \item \textbf{\textit{Wiping:}} Grasping a towel and wiping a designated area on the table surface.
\end{itemize}

To ensure a robust evaluation of generalization, we apply significant randomization to the initial state of the environment. For all tasks, the 6-DoF pose of primary objects (\eg, cubes, banana) and containers (\eg, box, drawer) is randomized. Specifically, their initial positions are sampled uniformly within an annular region 28 to 35 cm from the robot's base. Concurrently, their initial orientations around the vertical z-axis are randomized within intervals such as $[0, 2\pi]$, $[0, \pi]$, and $[-\pi/8, \pi/8]$.

\subsection{Results}
\subsubsection{Zero-shot Sim2Real Transfer}
To validate \our's ability to achieve zero-shot Sim2Real transfer, we conducted a key experiment where the policy is trained \textit{exclusively} on our simulated data and then deployed directly in the real world without any fine-tuning. \textit{We specifically analyze the performance of a policy trained on 100 demonstrations generated by \our (100 \our) and compare it to a policy trained on 50 real-world demonstrations (50 Real)}, as shown in Tab.~\ref{tab:exp}.

\subsubsection{Synergistic Effect of Real and Simulated Data}
To evaluate the benefit of combining real and simulated data, we compare policies trained on 50 real-world demonstrations (50 Real) against policies trained on a hybrid dataset (50 Real + 50 \our). The results, presented in Table \ref{tab:exp}, reveal a substantial and consistent performance boost across all evaluated methods. This highlights a powerful synergistic effect, where our high-fidelity synthetic data significantly enhances the utility of limited real-world data.

\subsubsection{Data Scaling and Efficiency Analysis}
We conduct a data scaling analysis to quantify the quality and sample efficiency of our simulated data. Using Diffusion Policy~\citep{chi2023diffusion}, we compare performance of policies trained on varying amounts of real data (50, 100 demonstrations) versus those trained exclusively on our simulated data (50 to 500 demonstrations). As shown in Fig.~\ref{fig:scaling}, the policy trained on just 200 of demonstrations by \our achieves performance comparable to one trained on 100 real-world demonstrations.

\subsubsection{Generalization to Challenging Settings}
A robust policy must generalize to unforeseen variations. We test this by evaluating performance in challenging real-world conditions, as shown in Fig.~\ref{fig:challenging_settings}. We compare three policies: one trained on 50 real demonstrations, a second on 50 simulated demonstrations with our holistic augmentations, and a third on their combination. As shown in Tab.~\ref{tab:exp1}, the policy trained with our augmented synthetic data significantly outperforms the one trained on real data alone, demonstrating superior generalization. Combining both data sources yields the best results, confirming that our augmentations effectively prepare the policy for real-world uncertainties.

\subsubsection{Sim2Real Transfer Fidelity}
\begin{figure}[t]
  \centering
  \includegraphics[width=0.7\linewidth]{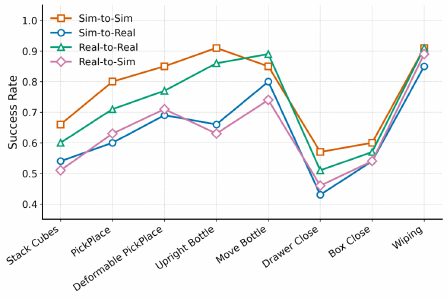}
  \caption{\textbf{Policy Performance Analysis Across Training and Evaluation Domains}. This chart compares the success rates of policies on eight tasks. We evaluate four distinct scenarios to analyze the domain gaps: policies are trained on data from our \our (`Sim') or from real-world demonstrations (`Real'), and then evaluated in both domains.}
  \label{fig:sim_real_p}
\end{figure}

To assess the core transfer capability of \our, we conduct a comprehensive cross-domain evaluation (Fig.~\ref{fig:sim_real_p}). The key finding is that policies trained exclusively in our simulator (Sim-to-Real) achieve success rates nearly identical to those of policies trained and evaluated entirely in the real world (Real-to-Real). This result validates our central claim of achieving true zero-shot Sim2Real transfer without any real-world fine-tuning. Furthermore, the strong performance in the Real-to-Sim scenario confirms that our simulation environment serves as a high-fidelity twin of reality, capable of reliably evaluating policies trained on real data.

\begin{table}[t]
 \centering
 % \scriptsize
\caption{\textbf{Performance of simulated data generation.} We report the data collection success rate (\textbf{Succ}) and average generation time in seconds (\textbf{Time}) per sample, measured on a single NVIDIA RTX 5060 Ti GPU.}
 \label{tab:data_collection}
 \setlength{\tabcolsep}{7pt} 
 \begin{tabular}{lcc|lcc}
 \toprule
 \rowcolor{headerpurple!60}
 \textbf{Task} & \textbf{Succ}  & \textbf{Time}  & \textbf{Task} & \textbf{Succ}  & \textbf{Time} \\
 \cmidrule(r){1-3} \cmidrule(l){4-6}
 \rowcolor{gray!10} 
 Stack Cubes & 0.94 & 8.9 & Move Bottle & 0.95  & 6.7  \\
 PickPlace & 0.98 & 8.6  & Drawer Close & 0.93  & 9.8  \\
 \rowcolor{gray!10} 
 Deformable PickPlace & 0.97  & 8.7  & Box Close & 0.95  & 10.3  \\
 Upright Bottle & 0.97 & 8.1  & Wiping & 1.00  & 6.2  \\
 \bottomrule
 \end{tabular}
\end{table}

\subsubsection{Data Generation Efficiency and Scalability}
In addition to demonstrating high success rates in simulation, our framework offers a significant advantage in data generation efficiency. As shown in Tab.~\ref{tab:data_collection}, \our generates over 10,000 demonstrations per day on a single NVIDIA RTX 5060 Ti GPU. In contrast, a full-time human operator can typically collect around 1,000 demonstrations per day via teleoperation. Our method thus offers a more than 10-fold increase in data collection throughput, providing a scalable and cost-effective alternative to manual data collection.

\subsection{Ablation study}
\subsubsection{Effect of Physics Estimation}
To isolate the contribution of our MLLM-driven Physics Estimation in Section.~\ref{sec:object_reconstruction}, we trained a policy on 50 demonstrations generated with default, physically implausible properties. As shown in Table~\ref{tab:exp}, removing physics estimation significantly degrades real-world performance, especially in contact-rich tasks. For instance, the success rate for Wiping plummets from 0.85 to 0.51, and for Deformable PickPlace, from 0.69 to 0.54. This occurs because the policy learns invalid interaction dynamics. This result underscores that accurate physics simulation is as critical as visual fidelity.

\subsubsection{Effect of Holistic Scene Augmentation}
To demonstrate that simple domain randomization is insufficient, we ablate our holistic augmentation strategy. We trained a policy on 50 demonstrations where only the 6-DoF pose of objects is randomized. As shown in Table~\ref{tab:exp} (50 \our w/o Holistic Scene Augmentation), this partial augmentation still results in a severe performance collapse. The policy overfits to the static camera view, lighting conditions, and deterministic action sequences. Therefore, our holistic approach, which randomizes the entire scene context, is indispensable for learning a truly robust and deployable policy.

%% file: Section/5_conclusion.tex
\section{CONCLUSIONS}
This work presents \our, a novel Real2Sim2Real framework tackling the core challenges in robot learning—the high cost of real-world data collection and the significant gap between simulation and reality. At its core is a hybrid scene representation merging the photorealism of 3DGS for static environments with the physical fidelity of mesh primitives for dynamic objects. We pioneer the use of a MLLM to deduce objects' physical properties and kinematic structures from vision, creating fully articulated assets automatically. Policies trained solely on our generated data achieve high success rates in diverse, real-world manipulation tasks, realizing zero-shot Sim2Real transfer. Furthermore, our method serves as a powerful data augmentation tool, substantially improving SOTA visuomotor policies when supplementing limited real-world data.

\medskip
\noindent
\textbf{Limitation}.
A primary limitation of RoboSimGS is the time-consuming and complex scene reconstruction pipeline. Although this manual step is a one-time effort per scene, it poses a significant bottleneck for large-scale deployment of our framework to new environments. Future work will focus on faster 3D reconstruction methods to streamline this setup.